\documentclass[conference]{IEEEtran}
\IEEEoverridecommandlockouts
\usepackage{cite}
\usepackage{amsmath,amssymb,amsfonts}
\usepackage{graphicx}
\usepackage{textcomp}
\usepackage{xcolor}

\usepackage{subfigure}

\usepackage{algorithm}
\usepackage{algpseudocode}
\usepackage{graphics}
\usepackage{epsfig}

\usepackage{multirow}
\usepackage{booktabs}

\usepackage{multicol}

\def\BibTeX{{\rm B\kern-.05em{\sc i\kern-.025em b}\kern-.08em
    T\kern-.1667em\lower.7ex\hbox{E}\kern-.125emX}}
\begin{document}


\title{Transferability of Adversarial Examples to Attack Cloud-based Image Classifier Service}

\author{\IEEEauthorblockN{Dou Goodman}
\IEEEauthorblockA{\textit{Baidu X-Lab} \\
Beijing, China \\
liu.yan@baidu.com}
}

\maketitle

\begin{abstract}
In recent years, Deep Learning(DL) techniques have been extensively deployed for computer vision tasks, particularly visual classification problems, where new algorithms reported to achieve or even surpass the human performance. While many recent works demonstrated that DL models are vulnerable to adversarial examples. Fortunately, generating adversarial examples usually requires white-box access to the victim model, and real-world cloud-based image classification services are more complex than white-box classifier,the architecture and parameters of DL models on cloud platforms cannot be obtained by the attacker. The attacker can only access the APIs opened by cloud platforms. Thus, keeping models in the cloud can usually give a (false) sense of security. In this paper, we mainly focus on studying the security of real-world cloud-based image classification services. Specifically, (1) We propose a novel attack method, Fast Featuremap Loss PGD (FFL-PGD) attack based on Substitution model, which achieves a high bypass rate with a very limited number of queries. Instead of millions of queries in previous studies, our method finds the adversarial examples using only two queries per image; and (2) we make the first attempt to conduct an extensive empirical study of black-box attacks against real-world cloud-based classification services. Through evaluations on four popular cloud platforms including Amazon, Google, Microsoft, Clarifai, we demonstrate that FFL-PGD attack has a success rate over 90\% among different classification services. (3) We discuss the possible defenses to address these security challenges in cloud-based classification services. Our defense technology is mainly divided into model training stage and image preprocessing stage.
\end{abstract}


\section{INTRODUCTION}

In recent years, Deep Learning(DL) techniques have been extensively deployed for computer vision tasks, particularly visual classification problems, where new algorithms reported to achieve or even surpass the human performance. Success of DL algorithms has led to an explosion in demand. To further broaden and simplify the use of DL algorithms, cloud-based services offered by Amazon\footnote{https://aws.amazon.com}, Google\footnote{https://cloud.google.com}, Microsoft\footnote{https://azure.microsoft.com}, Clarifai\footnote{https://www.clarifai.com/}, and others to offer various computer vision related services including image auto-classification, object identification and illegal image detection. Thus, users and companies can readily benefit from DL applications without having to train or host their own models.

\cite{szegedy2013intriguing} discovered an intriguing properties of DL models in the context of image classification for the first time. They showed that despite the state-of-the-art DL models are surprisingly susceptible to adversarial attacks in the form of small perturbations to images that remain (almost) imperceptible to human vision system. These perturbations are found by optimizing the input to maximize the prediction error and the images modified by these perturbations are called as adversarial example. The profound implications of these results triggered a wide interest of researchers in adversarial attacks and their defenses for deep learning in general.The initially involved computer vision task is image classification. For that, a variety of attacking methods have been proposed, such as L-BFGS of \cite{szegedy2013intriguing}, FGSM of \cite{goodfellow2014explaining}, PGD of \cite{madry2017towards},deepfool of \cite{moosavi2016deepfool} ,C\&W of \cite{Carlini2016Towards} and so on.

Fortunately, generating adversarial examples usually requires white-box access to the victim model, and real-world cloud-based image detection services are more complex than white-box classification and the architecture and parameters of DL models on cloud platforms cannot be obtained by the attacker. The attacker can only access the APIs opened by cloud platforms\cite{goodman2019cloud,goodman2020attacking}. Thus, keeping models in the cloud can usually give a (false) sense of security. Unfortunately, a lot of experiments have proved that attackers can successfully deceive cloud-based DL models without knowing the type, structure and parameters of the DL models\cite{goodman2019defcontransferability,goodman2019hitbtransferability}.

In general, in terms of applications, research of adversarial example attacks against cloud vision services can be grouped into three main categories: query-based attacks, transfer learning attacks and spatial transformation attacks. Query-based attacks are typical black-box attacks, attackers do not have the prior knowledge and get inner information of DL models through hundreds of thousands of queries to successfully generate an adversarial example \cite{Shokri2017Membership}. In \cite{ilyas2017query}, thousands of queries are required for low-resolution images. For high-resolution images, it still takes tens of thousands of times. For example, they achieves a 95.5\% success rate with a mean of 104342 queries to the black-box classifier. In a real attack, the cost of launching so many requests is very high.Transfer learning attacks are first examined by \cite{szegedy2013intriguing}, which study the transferability between different models trained over the same dataset. \cite{Liu2016Delving} propose novel ensemble-based approaches to generate adversarial example . Their approaches enable a large portion of targeted adversarial example to transfer among multiple models for the first time.However, transfer learning attacks have strong limitations, depending on the collection of enough open source models, but for example, there are not enough open source models for pornographic and violent image recognition.Spatial transformation attacks are simple and effective.\cite{Hosseini2017Google} found that adding an average of 14.25\% impulse noise is enough to deceive the Google’s Cloud Vision API.\cite{YuanStealthy} found 7 major categories of spatial transformation attacks to evade explicit content detection while still preserving their sexual appeal, even though the distortions and noise introduced are clearly observable to humans. 
To the best of our knowledge, no extensive empirical study has yet been conducted to black-box attacks and defences against real-world cloud-based image classification services. We summarize our main contributions as follows:
\begin{itemize}
	\item We propose a novel attack method, Fast Featuremap Loss PGD(FFL-PGD) attack based on Substitution model ,which achieves a high bypass rate with a very limited number of queries. Instead of millions of queries in previous studies, our method finds the adversarial examples using only one or twe of queries.
	\item We make the first attempt to conduct an extensive empirical study of black-box attacks against real-world cloud-based image classification services. Through evaluations on four popular cloud platforms including Amazon, Google, Microsoft, Clarifai, we demonstrate that our FFL-PGD attack has a success rate almost 90\% among different classification services.
	\item We discuss the possible defenses to address these security challenges in cloud-based classification services. Our protection technology is mainly divided into model training stage and image preprocessing stage.
\end{itemize}

\section{THREAT MODEL AND CRITERION}
\subsection{Threat Model}
In this paper, we assume that the attacker can only access the APIs opened by cloud platforms, and get inner information of DL models through limited queries to generate an adversarial example.Without any access to the training data, model, or any other prior knowledge,is a real black-box attack.
\subsection{Criterion and Evaluation}
The same with \cite{Li2019Adversarial},We choose top-1 misclassification as our criterion,which means that our attack is successful if the label with the highest probability generated by the neural networks differs from the correct label.We assume the original input is $O$,the adversarial example is $ADV$. For an RGB image $(m \times n \times 3)$, $(x,y,b)$ is a coordinate of an image for channel $b(0 \leqslant b \leqslant 2)$ at location $(x,y)$.We use Peak Signal to Noise Ratio (PSNR)\cite{Amer2002} to measure the quality of images. 
\begin{equation}
PSNR = 10log_{10} (MAX^2/MSE)
\end{equation}
where $MAX =255$, $MSE$ is the mean square error.
\begin{equation}
MSE = \frac{1}{mn*3}*\sum_{b=0}^2\sum_{i=1}^n\sum_{j=1}^m ||ADV(i,j,b)-O(i,j,b)||^2
\end{equation}
Usually, values for the PSNR are considered between 20 and 40 dB, (higher is better) \cite{amer2005fast}.We use structural similarity (SSIM) index to measure image similarity, the details of how to compute SSIM can be found in \cite{wang2004image}.Values for the SSIM are considered good between 0.5 and 1.0, (higher is better).
\section{BLACK-BOX ATTACK ALGORITHMS}
\subsection{Problem Definition}
A real-world cloud-based image classification service is a function $F(x) = y$ that accepts an input image $x$ and produces an output $y$. $F(.)$ assigns the label $C(x)=\arg \max_{i}F(x)_i$ to the input $x$. 

Original input is $O$, the adversarial example is $ADV$ and $\epsilon$ is the perturbation.

Adversarial example is defined as:

\begin{equation}
ADV=O+\epsilon
\end{equation}

We make a black-box  untargeted attack against real-world cloud-based classification services $F(x)$:

\begin{equation}
C(ADV)\not=C(O)
\end{equation}

We also assume that we are given a suitable loss function $L( \theta ,x,y)$,for instance the cross-entropy loss for a neural network. As usual,  $\theta \in \mathbb{R}^p$ is the set of model parameters.

\subsection{Fast Featuremap Loss PGD based on Substitution model}
\cite{Papernot2016Practical} proposed that the attacker can train a substituted model, which approaches the target model, and then generate adversarial examples on the substituted model. Their experiments showed that good transferability exists in adversarial examples.But the attack is not totally black-box. They have knowledge of the training data and test the attack with the same distributed data, and they upload the training data themselves and they know the distribution of training data \cite{Papernot2016Practical}\cite{Chen2017ZOO}\cite{Hayes2017Machine}. This leads us to propose the following strategy: 
\begin{enumerate}
	\item Substitute Model Training: the attacker queries the oracle with inputs selected by manual annotation to build a model $F{}'(x)$ approximating the oracle model $F(x)$ decision boundaries.
	\item Adversarial Sample Crafting: the attacker uses substitute network $F{}'(x)$  to craft adversarial samples, which are then misclassified by oracle $F(x)$ due to the transferability.We propose Fast Featuremap Loss PGD attack  to improve the success rate of transfer attack.
\end{enumerate}	

\begin{figure}[htbp] 
	\centering 
	\includegraphics[width=0.5\textwidth]{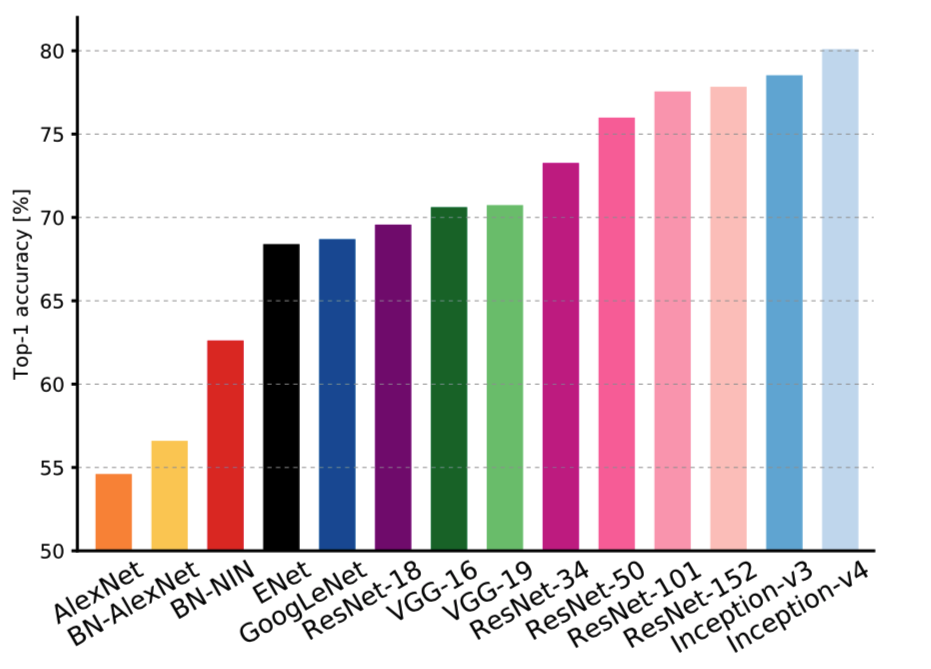}
	\caption{Top1 vs. network. Top-1 validation accuracies for top scoring single-model architectures \cite{Canziani2016An}.} 
	\label{VGG} 
\end{figure}

\subsubsection{Substitute Model Training Algorithm}  
We observe that a large number of machine vision tasks utilize feature networks as their backends. For examples, Faster-RCNN \cite{Ren2015Faster} and SSD \cite{Liu2016SSD} use the same VGG-16 \cite{simonyan2014very}. If we destroy the extracted features from the backend feature network, both of them will be influenced. 

We can choose one of AlexNet \cite{krizhevsky2012imagenet}, VGG \cite{simonyan2014very}and ResNet \cite{he2016deep}  which pretrained on ImageNet  as our substitute model. Better top-1 accuracy means stronger feature extraction capability. As can be seen from Figure 1, ResNet-152 has relatively good top-1 accuracy, so we choose ResNet-152 as our substitute model. We fix the parameters of the feature layer and train only the full connection layer of the last layer.

Our Substitute Model Training Algorithm is very simple, we train substitute model and generate adversarial example with the same images.

\begin{algorithm}[t]
	\caption{\textbf{- Substitute Model Training:} for oracle $F$, a substitute architecture $F{}'$, and an initial training set $S$.}
	\label{alg:substitute-training}
	\begin{algorithmic}[1]
		\Require $F{}'(x)$ ,$S$
		\State Define architecture $F{}'(x)$ 
		\State //Label the substitute training set with $F$
		\State $S \leftarrow \left\{ (x, F(x)) : x\in S \right\}$
		\State // Train $F{}'$ on $S$ to evaluate parameters $\theta_F{}'$
		\State $\theta_F{}' \leftarrow \mbox{train}(F{}',S)$
		\State \Return $\theta_F$
	\end{algorithmic}
\end {algorithm}

\subsubsection{Adversarial Sample Crafting Algorithm}
Previous work \cite{xie2017mitigating} has shown that one-step or multi-step attack algorithm such as $FGSM$ and $FGSM^k$, has better robustness in transfer attacks than $CW2$, which is based on optimization.$FGSM$ is an attack for an $\ell_{\infty}$ -bounded adversary and computes an adversarial example as:

\begin{equation}
x+\varepsilon \operatorname{sgn}\left(\nabla_{x} L(\theta, x, y)\right)
\end{equation}

A more powerful adversary is the multi-step variant $FGSM^k$, which is essentially projected gradient descent ($PGD$) on the negative loss function \cite{madry2017towards}:

\begin{equation}
x^{t+1}=\Pi_{x+\mathcal{S}}\left(x^{t}+\varepsilon \operatorname{sgn}\left(\nabla_{x} L(\theta, x, y)\right)\right)
\end{equation}

As usual, loss function $L( \theta ,x,y)$ is cross-entropy loss for a neural network. We propose Fast Featuremap Loss PGD attack which has a novel loss function to improve the success rate of transfer attack.The loss function $L$ is defined as:
\begin{equation}
L = class\_loss+\beta*FeatureMaps\_loss
\end{equation}

Where $\varepsilon$ and $\beta$ are the relative importance of each loss function. Next, we will introduce each component of the loss function in detail.

\textbf{Class Loss}
The core goal of generating adversarial example is to make the result of classification wrong. The first part of our loss function is class loss. Assuming that the Logits output of the classifier can be recorded as $Z(x)$, the output value corresponding to the classification label $i$ is $Z_{i}(x)$, and $t$ is the label of normal image. The greater the value of $Z_{i}(x)$, the greater the confidence that $x$ is recognized as $i$ by the classifier.
\begin{equation}
class\_loss (x) = \max(\max \{ Z(x)_i : i \ne t\} - Z(x)_t, -\kappa).
\end{equation}
$\kappa$ is a hyperparameter and $\kappa$ is a positive number, the greater the $\kappa$ is, the greater the confidence of the adversarial example is recognized as $\kappa$ by the classifier. \cite{Carlini2016Towards} discussed the performance of various class loss in detail in CW2 algorithm. We choose the class loss selected in CW2 algorithm. The empirical value of $\kappa$ is 200.
\textbf{FeatureMap Loss}
Only class loss and distance loss can be used to generate adversarial example, which is what the L-BFGS algorithm of \cite{szegedy2013intriguing} does. However, as a black-box attack, we have no knowledge of the parameters and structure of the attacked model. We can only generate adversarial example through the known substitute model of white-box attack, and then attack target  model. The success rate of the attack depends entirely on the similarity between the substitute model and the attacked model. We introduce Feature Maps loss, which is the output of the last convolution layer of the substitute model, representing the highest level of semantic features of the convolution layer after feature extraction layer by layer $L_n$.

\begin{figure*}[htbp] 
	\centering 
	\includegraphics[width=0.8\textwidth]{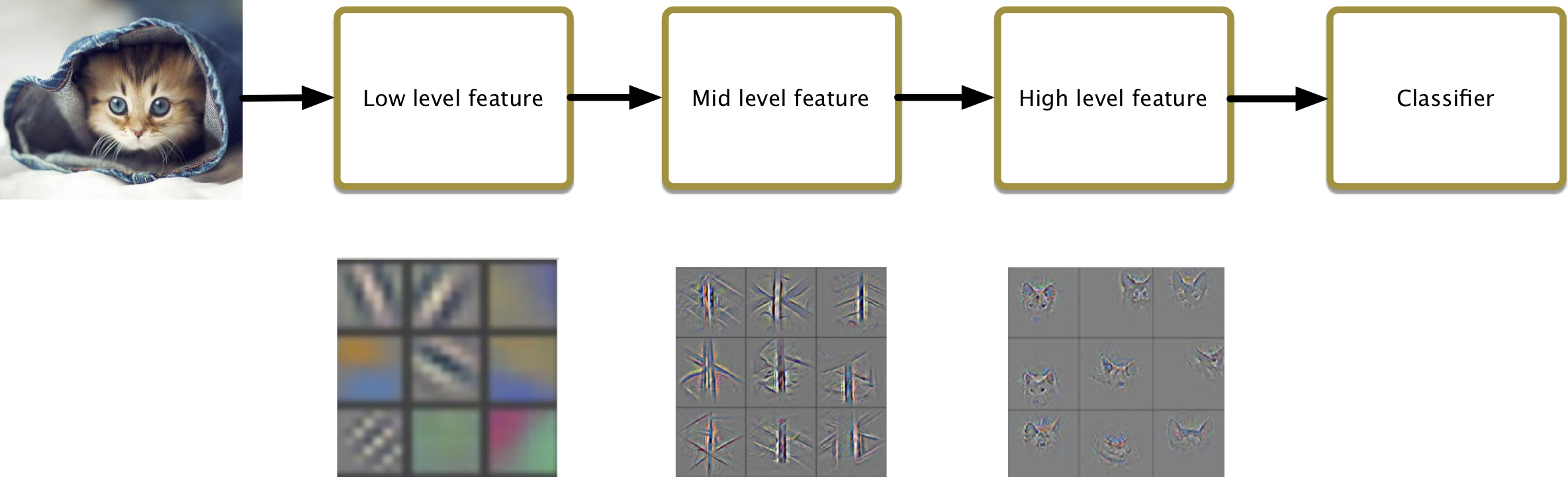}
	\caption{Illustration of cat recognition, the first convolution layer mainly recognizes low level features such as edges and lines. In the last convolution layer, it recognizes high level features such as eyes and nose.} 
	\label{cat_conv} 
\end{figure*}

We assume the original input is $O$, the adversarial example is $ADV$, and the featuremap loss is:
\begin{equation}
FeatureMap\_loss(ADV,O)=\|L_n(ADV)  -  L_n(O)  \|_2
\end{equation}
{zeiler2014visualizing}  visualizes the differences in the features extracted from each convolution layer. In cat recognition, for example, Figure~\ref{cat_conv} ,the first convolution layer mainly recognizes low level features such as edges and lines. In the last convolution layer, it recognizes high level features such as eyes and nose. In machine vision tasks, convolution layer is widely used for automatic feature extraction. And a large number of models are based on the common VGG, ResNet pre-training parameters on ImageNet and  finetuned the weights on the their own dataset. We assume that the feature extraction part of the main stream cloud-based image classification services are based on common open source models as VGG or others. The larger the feature Maps loss of the adversarial example and the original image, the greater the difference in the semantic level. We define the hyperparameter $\beta$. The larger the $\beta$, the better the transferability.

\section{EXPERIMENTAL EVALUATION}
\subsection{Datasets and Preprocessing}
100 cat images  and 100 other animal images  are selected from the ImageNet val set. Because VGG19 and Resnet50 both accept input images of size $224 \times 224 \times 3$, every input image is clipped to the size of $224 \times 224 \times 3$, where 3 is the number of RGB channels. The RGB value of the image is between 0 and 255. We use these 100 images of cats as original images to generate adversarial examples and make a black-box untargeted attack against real-world cloud-based image classification services. We choose top-1 misclassification as our criterion, which means that our attack is successful if the label with the highest probability generated by the cloud-based image classification service differs from the correct label "cat". We count the number of top-1 misclassification to calculate the escape rate.

\begin{table}[h]
	\caption{Correct label by cloud APIs}
	\label{tab:o}
	\centering
	\begin{tabular}{cccc}
		\toprule
		Platforms & Cat Images& Other Animal Images& All Images \\
		\midrule
		Amazon & 99/100 & 98/100 & 197/200\\
		Google & 97/100 & 100/100 &197/200 \\
		Microsoft & 58/100 & 98/100& 156/200 \\
		Clarifai & 97/100 & 98/100 &195/200 \\
		\bottomrule
	\end{tabular}
\end{table}



According to Table \ref{tab:o}, we can learn that Amazon and Google, which label 98.5\% of all  images correctly, have done a better job than other cloud platforms. 

\subsection{Fast Featuremap Loss PGD based on Substitution model}
We choose ResNet-152 as our substitute model, fix the parameters of the feature layer and train only the full connection layer of the last layer. We launched PGD and FFL-PGD attacks against our substitute model to generate adversarial examples. PGD and FFL-PGD share the same hyperparameter of step size $\varepsilon$ ,while the hyperparameter $\beta$ of FFL-PGD set to $0.1$.

\begin{figure*}[htbp]
	\subfigure[]{
		\label{fig:ffl:a}
		\begin{minipage}[t]{0.3\linewidth}
			\centering
			\includegraphics[width=2in]{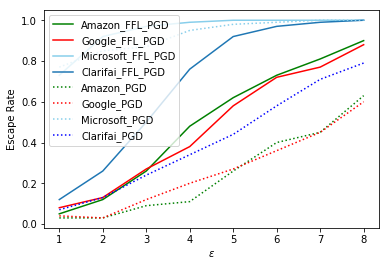}
		\end{minipage}
	}%
	\subfigure[]{
		\label{fig:ffl:b}
		\begin{minipage}[t]{0.3\linewidth}
			\centering
			\includegraphics[width=2in]{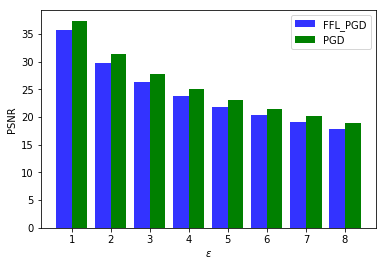}
		\end{minipage}
	}%
	\subfigure[]{
	\label{fig:ffl:c}
	\begin{minipage}[t]{0.3\linewidth}
		\centering
		\includegraphics[width=2in]{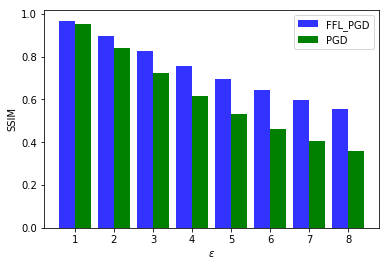}
	\end{minipage}
	}%
	\centering
	\caption{In \ref{fig:ffl:a}, we increase step size $\epsilon$ from 1 to 8, the figure records the escape rates of PGD and FFL-PGD attacks against cloud-based image classification services under different $\epsilon$. In \ref{fig:ffl:b}, the figure records the PSNR of PGD and FFL-PGD attacks and In \ref{fig:ffl:c}, the figure records the SSIM of PGD and FFL-PGD attacks. }
\end{figure*}

The escape rates of PGD and FFL-PGD attacks are shown in Figure \ref{fig:ffl:a}. From Figure \ref{fig:ffl:a}, we know that the cloud-based image classification services of Amazon, Google, Microsoft and Clarifai are vulnerable to PGD and FFL-PGD attacks . Step size $\epsilon$ controls the escape rate. Increasing this parameter can improve the escape rate.

When $\epsilon$ is the same, FFL-PGD has a higher escape rate than PGD. It can be seen that the FeatureMap Loss which added to the loss function is beneficial to improve the escape rate, that is, to improve the robustness of transfer attacks against different cloud-based image classification services.

From Figure \ref{fig:ffl:b} , we know that PGD has a higher PSNR ,which is considered as better image quality .But both of them higher than 20dB when $\epsilon$ from 1 to 8, which means both of them are considered acceptable for image quality.In addition, we can find that increasing $\epsilon$ will lead to image quality degradation.

From Figure \ref{fig:ffl:c} , we know that FFL-PGD has a higher SSIM ,which is considered as better image similarity .

FFL-PGD attack has a success rate over 90\% among different cloud-based image classification services and is considered acceptable for image quality and similarity using only two queries per image.

\cite{Liu2016Delving} adopted an ensemble-based model to improve transferability of attack and successfully attack Clarifai. We used their methods to attack the same cloud platforms and train our ensemble-based model with AlexNet, VGG-19, ResNet-50, ResNet-110 and ResNet-152. 



\begin{table*}[!tp]
	\caption{PSNR And SSIM of Ensemble-based Mode Attack}
	\label{tab:Ensemble}
	\centering
	\begin{tabular}{cccccc}
		\toprule
		The number of iteration &10&20 & 50 &100 & 200 \\
		\midrule
		PSNR& 26.53 &27.13 & 33.56&37.88&42.49\\
		SSIM&  0.60& 0.58 &0.64&0.70&0.77\\
		\bottomrule
	\end{tabular}
\end{table*}

According to Table \ref{tab:Ensemble}, we can learn that increasing the number of iteration can increase PSNR and SSIM under ensemble-based model attack, which means better image  quality  and similarity .

We can infer that when the number of iterations continues to increase, the perturbation $l_2$ value of adversarial examples decreases and the PSNR increases. Although the image quality can be improved, adversarial examples are over fitting  the model , and the transferability decreases in the face of the pretreatment of cloud services. In \cite{Xie2018Mitigating}, they take advantage of the weakness of iteration-based white-box attack, and use the pre-processing steps of random scaling and translation to defense the adversarial examples.

\section{DISCUSSION}
\subsection{Effect of Attacks}

Our research shows that FFL-PGD attack can reduce the accuracy of mainstream image classification services in varying degrees. To make matters worse, for any image classification service, we can find a way that can be almost 90\% bypassed.


\subsection{Defenses}
Defense adversarial examples is a huge system engineering, involving at least two stages: model training and image preprocessing.
\subsubsection{Model Training}
\cite{goodfellow2014explaining} proposed adversarial training to  improve the robustness of deep learning model. Retraining the model with new training data may be very helpful. Adversarial training included adversarial examples in the training stage and generated adversarial examples in every step of training and inject them into the training set. On the other hand, we can also generate adversarial samples offline, the size of adversarial samples is equal to the original data set, and then retrain the model. We have developed AdvBox\cite{goodman2020advbox}\footnote{https://github.com/advboxes/AdvBox}, which is convenient for developers to generate adversarial samples quickly. 

\subsubsection{Image Preprocessing}

\cite{Dziugaite2016} evaluated the effect of JPG compression on the classification of adversarial images and their experiments demonstrate that JPG compression can reverse small adversarial perturbations. However, if the adversarial perturbations are larger, JPG compression does not reverse the adversarial perturbation. \cite{xie2017mitigating} proposed a randomization-based mechanism to mitigate adversarial effects  and their experimental results show that adversarial examples rarely transfer between different randomization patterns, especially for iterative attacks. In addition, the proposed randomization layers are compatible to different network structures and adversarial defense methods, which can serve as a basic module for defense against adversarial examples. 


Although all the above efforts can only solve some problems, chatting is better than nothing.

\section{RELATED WORK}
Previous works mainly study the security and privacy in DL models via white-box mode \cite{szegedy2013intriguing} \cite{goodfellow2014explaining} \cite{madry2017towards} \cite{moosavi2016deepfool}. In the white-box model, the attacker can obtain the adversarial examples quickly and accurately. However, it is difficult for the attacker to know the inner parameters of models in the real world, so  researchers have launched some black-box attacks on  DL models recently. In general, in terms of applications, research of adversarial example attacks against cloud vision services can be grouped into three main categories: query-based  attacks, transfer learning  attacks and spatial transformation  attacks.

Query-based attacks are  typical black-box attacks, attackers do not have the prior knowledge and get inner information of DL models through hundreds of thousands of queries to successfully generate an adversarial example \cite{Shokri2017Membership}.In \cite{ilyas2017query}, thousands of queries are required for low-resolution images. For high-resolution images, it still takes tens of thousands of times. But attacking an image requires thousands of queries, which is not operable in actual attacks of real-world cloud-based image classification services.

In order to reduce the number of queries,\cite{Papernot2016Practical} attack strategy consists in training a local model to substitute for the target DL models, using inputs synthetically generated by an adversary and labeled by the target DL models. They have knowledge of the training data and test the attack with the same distributed data, and they upload the training data themselves and they know the distribution of training data \cite{Papernot2016Practical}\cite{Chen2017ZOO}\cite{Hayes2017Machine} . 

Transfer learning attacks are first examined by \cite{szegedy2013intriguing}, which study the transferability between different models trained over the same dataset. \cite{Liu2016Delving} propose novel ensemble-based approaches to generate adversarial example . Their approaches enable a large portion of targeted adversarial example to transfer among multiple models for the first time.

Spatial transformation  attacks are very interesting, \cite{Hosseini2017Google} evaluate the robustness of Google Cloud Vision API to input perturbation, they show  that adding an average of 14.25\% impulse noise is enough to deceive the API and when a noise filter is applied on input images, the API generates mostly the same outputs for restored images as for original images.

\cite{YuanStealthy} report the first systematic study on the real-world adversarial images and their use in online illicit promotions. They categorize their techniques into 7 major categories, such as color manipulation, rotation, noising and blurring. \cite{Li2019Adversarial} make the first attempt to conduct an extensive empirical study of black-box attacks against real-world cloud-based image detectors such as  violence, politician and pornography detection.

Our FFL-PGD attack based on Substitution model can be classified as a combination of query-based attack and transfer learning attack.

\section{CONCLUSION AND FUTURE WORK}
In this paper, (1) We propose a novel attack method, Fast Featuremap Loss PGD (FFL-PGD) attack based on Substitution model, which achieves a high bypass rate with a very limited number of queries. Instead of millions of queries in previous studies, our method finds the adversarial examples using only two queries per image; and (2) we make the first attempt to conduct an extensive empirical study of black-box attacks against real-world cloud-based classification services. Through evaluations on four popular cloud platforms including Amazon, Google, Microsoft, Clarifai, we demonstrate that FFL-PGD attack has a success rate almost 90\% among different classification services. (3) We discuss the possible defenses to address these security challenges in cloud-based classification services. Our defense technology is mainly divided into model training stage and image preprocessing stage.
In the future, we aim to explore the space of adversarial examples with less perturbation in black-box and attempt to study target attack using FFL-PGD attack. On the other hand, we will focus on the defense in the cloud environment, so that AI services in the cloud environment away from cybercrime. We hope cloud service providers will not continue to forget this battlefield.

\bibliographystyle{IEEEtran}
\bibliography{IEEEexample}

\end{document}